\documentclass{article}
\usepackage{spconf,amsmath,graphicx, dsfont, amsfonts, hyperref}

\title{Bilinear Representation for Language-Based Image Editing using Conditional Generative Adversarial Networks}
\name{Xiaofeng Mao \qquad Yuefeng Chen \qquad Yuhong Li \qquad Tao Xiong \qquad Yuan He \qquad Hui Xue}

\address{Alibaba Group, China\\
        \{mxf164419, yuefeng.chenyf, daniel.lyh, weilue.xt, heyuan.hy, hui.xueh\}@alibaba-inc.com}
\begin{document}
%
\maketitle
\begin{abstract}

The task of Language-Based Image Editing (LBIE) aims at generating a target image by editing the source image based on the given language description. The main challenge of LBIE is to disentangle the semantics in image and text and then combine them to generate realistic images. Therefore, the editing performance is heavily dependent on the learned representation. In this work, conditional generative adversarial network (cGAN) is utilized for LBIE. We find that existing conditioning methods in cGAN lack of representation power as they cannot learn the second-order correlation between two conditioning vectors. To solve this problem, we propose an improved conditional layer named Bilinear Residual Layer (BRL) to learning more powerful representations for LBIE task. Qualitative and quantitative comparisons demonstrate that our method can generate images with higher quality when compared to previous LBIE techniques.

\end{abstract}
\begin{keywords}    
Generative adversarial networks, Bilinear, Language-based image editing
\end{keywords}
\section{Introduction}
\label{sec:intro}

\begin{figure}
\begin{center}
\includegraphics[width=1\linewidth]{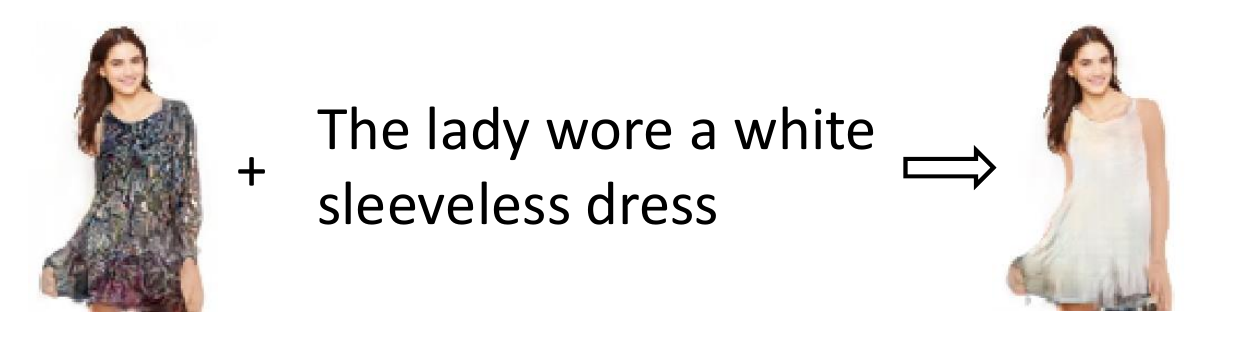}
\end{center}
   \vspace{-2.5em}
   \caption{LBIE for fashion generation.}
\label{fig:intro}
\vspace{-1.5em}
\end{figure}

The task of Language-Based Image Editing (LBIE) aims at manipulating a source image semantically to match the given description well. LBIE has seen applications to domains as diverse as Computer-Aided Design (CAD), Fashion Generation and Virtual Reality (VR)~\cite{chen2017language}. As illustrated in Fig~\ref{fig:intro}, using LBIE technique, one can automatically modify the color, texture or style for a given design drawing by language instructions instead of the traditional complex processes.


Nevertheless, LBIE is still challenging due to the following two difficulties: i). the model should find the areas in image which are relevant to the given text description; ii). the relations of disentangled semantics in image and text description should be learned for a better generation of realistic image. 
To tackle these problems, several methods have been proposed~\cite{chen2017language,zhu2017your,manjunatha2018learning, gunel2018language, dong2017semantic}, and most of them utilize the generative models, e.g., GANs~\cite{goodfellow2014generative}. ~\cite{chen2017language,zhu2017your} divide LBIE into two subtasks: language-based image segmentation and image generation. Specifically, Zhu et al. \cite{zhu2017your} performes LBIE to ``redress'' the person with the given outfit description, while at the same time keeping the wearer and his posture or expression unchanged. They use a two stages GAN that outputs a semantic segmentation map as intermediate step, which is further used to render the final image with precise regions and textures at the second step.
Some other approaches~\cite{manjunatha2018learning, gunel2018language, dong2017semantic} can achieve LBIE without any segmentation map or explicit spatial constraints by adversarially train a conditional GAN~\cite{mirza2014conditional}. Among them, \cite{dong2017semantic} is the seminal work and it uses concatenation operation to condition the image generation process with text embeddings. \cite{manjunatha2018learning,gunel2018language} follow up this framework, and replace the concatenation operation with Feature-wise Linear Modulation (FiLM), which is a more efficient and powerful method as a generalization of concatenation.


In this work, we first theoretically analyse these works which edit the image based on fused visual-text representations using different conditioning methods. We found that all these conditioning methods can be modeled by a universal form of bilinear transformation based on~\cite{dumoulin2018}.
However, all these methods are lack of representation power as they cannot learn the second-order correlation between two conditioning embeddings. To solve this problem, we present an improved conditoning method named Bilinear Residual Layer which can strike a happy compromise between representation effectiveness and model efficiency.
We have both theoretically and experimentally proved that the Bilinear Residual Layer can provide richer representations than previous approaches.
Quantitative and qualitative results on Caltech-200 bird \cite{WahCUB_200_2011}, Oxford-102 flower \cite{Nilsback08} and Fashion Synthesis datasets \cite{zhu2017your} suggest that our approach can generate images with higher quality when compared to previous LBIE techniques.

\section{METHOD}
\label{sec:method}

In this section, we first theoretically analyse existing conditioning methods in cGANs. Then an improved conditional layer called Bilinear Residual Layer (BRL) is proposed in Sec~\ref{subsection:brl}. Finally, we introduce overall framework in Sec~\ref{subsection:overall}.


\subsection{Overview of conditioning methods}
\label{subsec:overview}
Conditioning is a general-purpose operation and can be used for different tasks, e.g., conditional image generation \cite{huang2018synthesis, chen2018frontal} and cross-modality distillation \cite{roheda2018cross}. The most commonly used approach in conditional GANs is concatenation.
Formally, denote $I_{f} \in \mathbb{R}^{D}$ and $I_{c}\in \mathbb{R}^{{D}'}$ as the output of previous layer and conditioning feature respectively, where $D$ and ${D}'$ are the dimensionality of features. The concated representation $\left [ I_{f}\ I_{c} \right ] \in \mathbb{R}^{D+{D}'} $ can be further encoded by a matrix $W=\left [ W_{f}; W_{c} \right ]$, $W_{f}\in \mathbb{R}^{D\times O}$ and $W_{c}\in \mathbb{R}^{{D}'\times O}$ are the corresponding weights for $I_{f}$ and $I_{c}$. $O$ is the output dimension. Formally, we can get the following transformation:
\begin{equation}
    I_{o}=\left [ I_{f}\ I_{c} \right ]\left [ W_{f}; W_{c} \right ]= I_{f}W_{f}+I_{c}W_{c}
    \label{eq:1}
\end{equation}
where $I_{o}$ is the output tensor. Equation~\ref{eq:1} suggests that concatenation based conditioning method amounts to adding a feature-wise bias on the unconditional output $I_{f}W_{f}$. Therefore, some other approaches~\cite{vanwavenet,van2016conditional} suggest to add conditional bias directly instead of concatenation. 

Recently, some works~\cite{wang2017residual} have validated that deep models could mimic the human attention mechanism by gating each feature using a value between 0 and 1. Inspired by this, Perez et al. \cite{perez2018film} proposes a more general conditioning method named feature-wise linear modulation (FiLM), which rescales the features by adding multiplicative interactions:
\begin{equation}
    I_{o}=(I_{f}W_{f})\odot (I_{c}\overline W_{c} )+I_{c}W_{c}
    \label{eq:2}
\end{equation}
$\overline W_{c}\in \mathbb{R}^{{D}'\times O}$ is the weight for learning rescaling coefficients. From this formulation, we can conclude that concatenation is a special case of FiLM when $I_{c}\overline W_{c}=\mathds{1}$, where $\mathds{1}$ is a matrix of ones. FiLM has shown its superiority over conventional concatenation method and has been widely applied to the multimodal interaction.

\renewcommand{\thefootnote}{\fnsymbol{footnote}}

However, concatenation and FiLM only apply a linear transformation between the input and conditional features. In this work, we go a step further and generalize these linear methods to the more powerful bilinear version, which can provide richer representations than linear models by learning the second-order interaction. 
In bilinear model, the $i$th feature in output $I_{o}$ can be calculated as
\begin{equation}
    I_{o_{i}}=I_{f}\mathbf{W_{i}}I_{c}^{T}
    \label{eq:3}
\end{equation}
$\mathbf{W_{i}} \in \mathbb{R}^{D\times{D}'}$ is a weight matrix for the output feature $I_{o_{i}}$. Interestingly, we have found FiLM can be presented by bilinear transformations. Denote the weights corresponding to the $i$th output feature in $W_{f}$, $\overline W_{c} $ and $W_{c} $ as $w_{f_{i}}$, $\overline w_{c_{i}}$ and $w_{c_{i}}$. The FiLM transformation for $I_{o_{i}}$ = $(I_{f}w_{f_{i}})(I_{c}\overline w_{c_{i}}) + I_{c}w_{c_{i}}$ can be represented by
\begin{equation}
    I_{f}(\underbrace{w_{f_{i}}\overline w_{c_{i}}^{T}+\mathbf{{W_{i}}}'}_{\mathbf{W_{i}}})I_{c}^{T}
\end{equation}
where $I_{f}\mathbf{{W_{i}}}'=w_{c_{i}}^{T}$,  $\mathbf{{W_{i}}}'$ can be constructed by randomly choosing a nonzero element $I_{f_{k}}$ in $I_{f}$, we have 
\begin{equation}
    \mathbf{{W_{i}}}'=  \begin{bmatrix}
 0&  0&  0&  \cdots& \\ 
 \vdots&   \vdots&   &  & \\ 
 \frac{w_{c_{i1}}}{I_{f_{k}}}&  \frac{w_{c_{i2}}}{I_{f_{k}}}&  \cdots&  \frac{w_{c_{i{D}'}}}{I_{f_{k}}}& \\ 
 \vdots&  \vdots&  &  & \\ 
 0&  0&  0&  \cdots& 
            \end{bmatrix}
\end{equation}
where elements in the $\mathbf{{W_{i}}}'$ are 0 except for the $k$th row. Obviously, the rank of matrix $w_{f_{i}}\overline w_{c_{i}}^{T}$ and $\mathbf{{W_{i}}}'$ are both 1. So we have $Rank(\mathbf{W_{i}})\leq 2$\footnote{Properties of rank: \href{https://en.wikipedia.org/wiki/Rank_(linear_algebra)}{https://en.wikipedia.org/wiki/Rank\_(linear\_algebra)}}. The constructed formulation indicates that FiLM is equivalent to bilinear transformation with transformation matrix $\mathbf{W_{i}}$ is sparse and has the rank no greater than 2. From a theoretical perspective, it illustrates that bilinear transformations can provide more fine-grained conditioning representations than the concatenation and FiLM.

\subsection{Bilinear Residual Layer}
\label{subsection:brl}
\begin{figure}
\begin{center}
\includegraphics[width=1\linewidth]{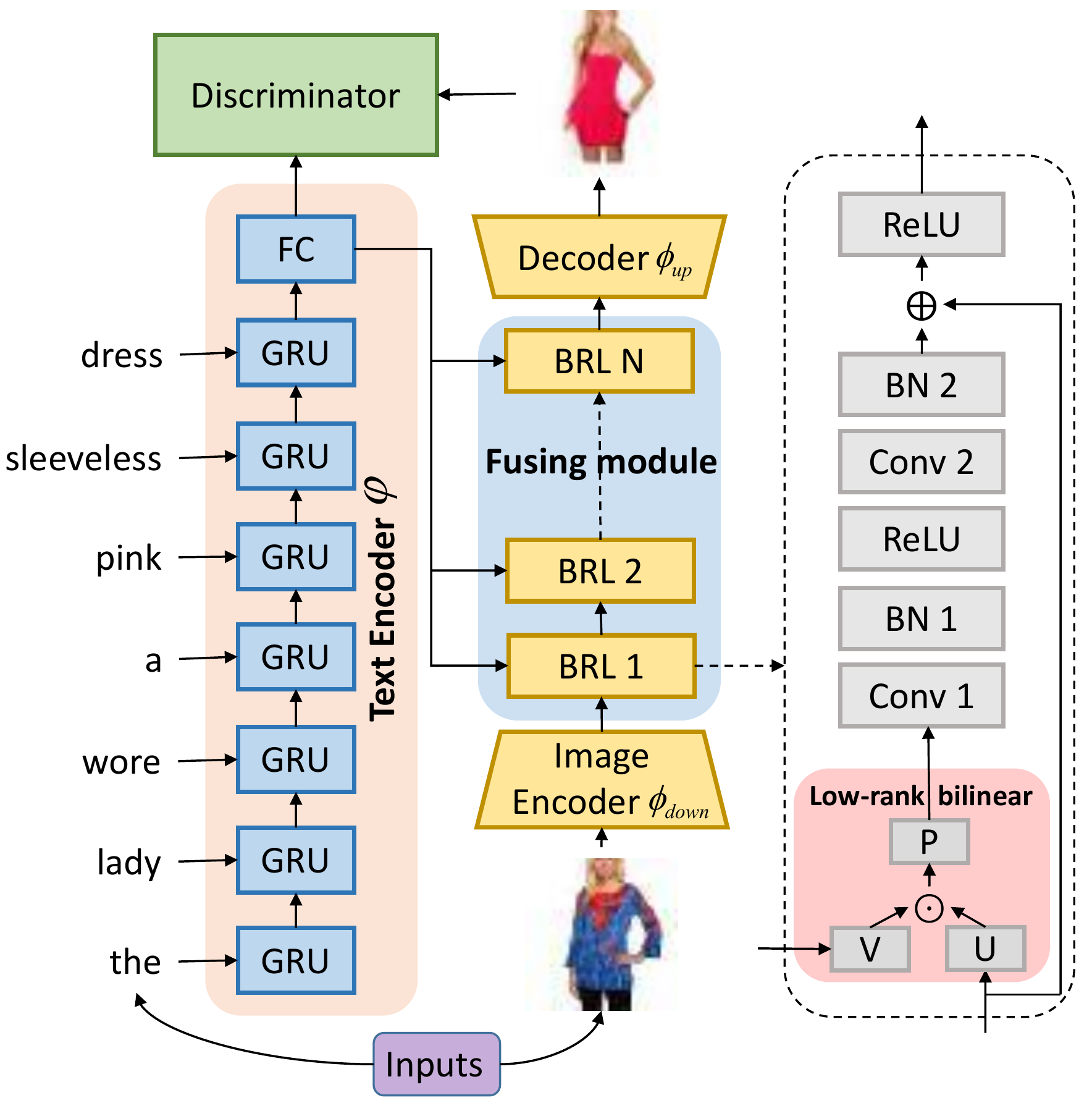}
\end{center}
   \vspace{-1em}
   \caption{Overview of our network architecture. Detail of our bilinear residual layer is presented in the dashed box.}
\label{fig:architecture}
\vspace{-1em}
\end{figure}

We propose Bilinear Residual Layer (BRL) for learning conditional bilinear representations as illustrated in dashed box of Fig~\ref{fig:architecture}. Similar to FiLM, we add shortcuts to guarantee the model's capability to learn identical mapping. As a consequence, our bilinear residual layer can automaticly decide whether or not the model needs to incorporate the conditioning information in the later layers. 

However, the representational power of bilinear features comes with the cost of very high-dimensional model parameters , which require substantial
computing and large quantities of training data to fit~\cite{kong2017low}. For example, the dimensionality of $\mathbf{{W}}$ is $|D\times{D}'\times O|$ which is cubical expansion.
To reduce the dimensionality of model parameters, our approach adopts a low-rank bilinear method \cite{kim2016hadamard} to reduce the rank of $\mathbf{W_{i}}$. Based on this idea, $I_{o_{i}}$ can be rewritten as follows:
\begin{equation}
    I_{o_{i}}=I_{f}\mathbf{W_{i}}I_{c}^{T}=I_{f}\mathbf{U_{i}}\mathbf{V_{i}}^{T}I_{c}^{T}=I_{f}\mathbf{U_{i}}\odot I_{c}\mathbf{V_{i}}
\end{equation}
where $\mathbf{U_{i}} \in \mathbb{R}^{D\times d}$ and $\mathbf{V_{i}} \in \mathbb{R}^{{D}'\times d}$ are the decomposed submatrices and they restrict the rank of $\mathbf{W_{i}}$ to be at most $d\leq \min (D,{D}')$. Then the final feature vector $I_{o}$ can be projected by $\mathbf{P}\in \mathbb{R}^{O\times d}$ as follows:
\begin{equation}
    I_{o}=\mathbf{P}(I_{f}\mathbf{U}\odot I_{c}\mathbf{V})
\end{equation}

Moreover, Our bilinear residual layer is a general condition layer, and it is applicable not only for LBIE, but also for other conditional models or applications, e.g., text-to-image generation \cite{zhang2017stackgan}. In following sections, we will present the overall framework of our work and we denote the bilinear residual layer as $\mathcal{F}$ for convenience. 

\subsection{Overall framework}\label{subsection:overall}
We follow the work of Dong et al.~\cite{dong2017semantic} which utilizes the cGAN to learn the target mapping conditioning based on image and text description. As shown in Fig \ref{fig:architecture}, the network consists of a generator $G$ and a discriminator $D$. The generator has three modules: encoding module, fusing module and decoding module. Encoding module contains pre-trained encoders $\varphi$ and $\phi_{down}$, and they are used to extract text and image features respectively. We adopt the procedure in \cite{kiros2014unifying} to pre-train the text encoder $\varphi$ and use parameters of \emph{conv}1-4 layers in VGG16 as the feature extractor $\phi_{down}$ for image. The text and image features are then fed in the following fusing module, which can be seen as a conditioning layer to compromise the semantics of multiple modalities. The final decoding module $\phi_{up}$ upsamples the fused feature to a high-resolution images. Finally, the discriminator is a classifier which takes the generated image and text embeddings as input and output the probability whether the description matches the image.

Formally, given an original image-text pair $ \textless x,t \textgreater$, $t$ is the text matching with the image $x$. Suppose that we use description text $\hat{t}$ to manipulate the image $x$, typically $\hat{t}$ is a text relevant to $x$. The generator can transform the image according to text embedding $\varphi (\hat{t})$ and output
\begin{equation}
    G(x,\varphi (\hat{t}))=\phi_{up}(\mathcal{F}(\phi_{down}(x),\varphi(\hat{t})))
\end{equation}
the discriminator $D$ is trained to distinguish semantically differentiated image-text pairs. To this end, we need to take a mismatching text $\overline{t}$ as negative sample. Original pair $\textless x,t \textgreater$, current editing pair $\textless x,\hat{t} \textgreater$ and negative pair $\textless x,\overline{t} \textgreater$ are fed into discriminator $D$ to minimizing
\begin{equation}
\begin{split}
    \mathcal{L}_{D}&=\mathbb{E}_{(x,\overline{t})\sim p_{data}}\left [ D(x,\varphi (\overline{t}))^{2} \right ]\\
    &+\mathbb{E}_{(x,t)\sim p_{data}}\left [ (D(x,\varphi (t))-1)^2 \right ]\\
    &+\mathbb{E}_{(x,\hat{t})\sim p_{data}}\left [ D(G(x,\varphi (\hat{t})),\varphi (\hat{t}))^{2} \right ]
\end{split}
\end{equation}
here the objective of the first and second terms is to classify negative and original real-world image-text pairs. The third term makes $D$ to identify the synthesized image with its editing text as mismatching as possible. Alternately to the training of $D$, the generator $G$ is trained to generate more semantically similar images with the editing text $\hat{t}$:
\begin{equation}
    \mathcal{L}_{G}=\mathbb{E}_{(x,\hat{t})\sim p_{data}}\left [ (D(G(x,\varphi (\hat{t})),\varphi (\hat{t}))-1)^{2} \right ]
\end{equation}
In this work, $\overline{t}$ and $\hat{t}$ are selected from the text descriptions of other images in the dataset.

\begin{figure*}
\begin{center}
\includegraphics[width=1\linewidth]{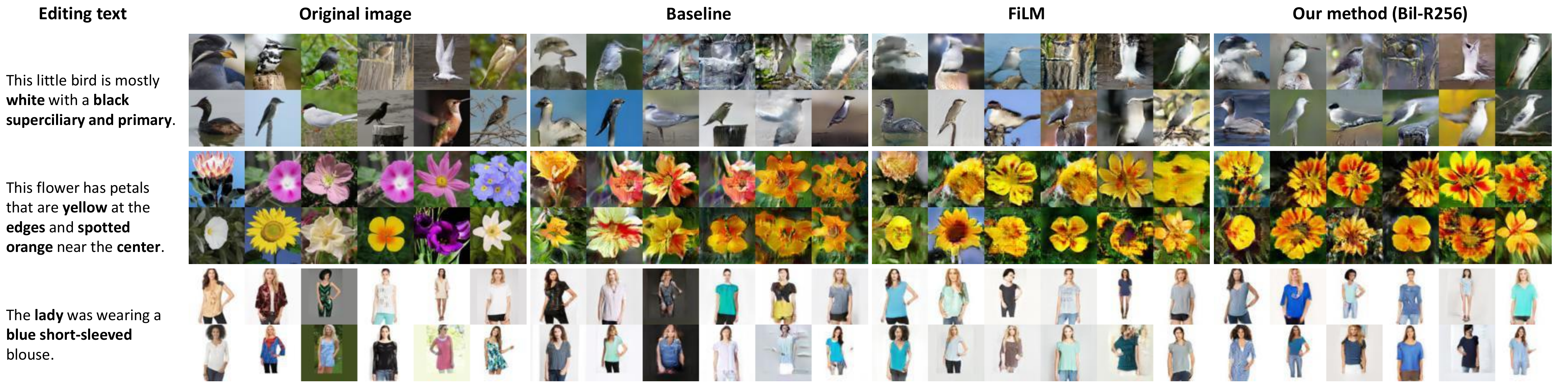}
\end{center}
   \vspace{-1.3em}
   \caption{Qualitative comparisons.}
\label{fig:exp}
\vspace{-0.8em}
\end{figure*}

\section{Experiments}
\label{sec:experiments}
We conduct experiments on Caltech-200 bird dataset \cite{WahCUB_200_2011}, Oxford-102 flower dataset \cite{Nilsback08} and Fashion Synthesis dataset \cite{zhu2017your}. The bird dataset has 11,788 images with 200 classes of birds. We split it to 160 training classes and 40 testing classes. The flower dataset has 8,189 images with 102 classes of flowers, and we split it to 82 training classes and 20 testing classes. The fashion dataset has much more classes with 78,979 images totally. We choose 3200 classes from 4119 for training and the rest for testing.
\subsection{Implementation details}
The source code has been released\footnote{\href{https://github.com/vtddggg/BilinearGAN_for_LBIE}{https://github.com/vtddggg/BilinearGAN\_for\_LBIE}}. Our encoder $\varphi$ for text descriptions is a recurrent network. Given the pair of image and text $\textless x, t \textgreater$, the method in~\cite{kiros2014unifying} is used to pre-train the text encoder to minimize the pair-wise ranking loss. This pre-trained text encoder encodes the text description $t$ into visual-semantic text representation $\varphi(t)$, which will be further used in the adversarial training process as detailed in Sec \ref{subsection:overall}.


For image encoder, it receives images with size of 64$\times$64 as input and output features with dimension of 16$\times$16$\times$512. Text encoder encodes descriptions to the text embeddings with dimensionality of 128. Our fusing module consists of 4 (i.e., N in the Fig.~\ref{fig:architecture}) bilinear residual layers. To implement the low-rank bilinear method, we duplicate the text embeddings to be of dimension 16$\times$16$\times$128, so as to keep the same spatial size with image features. Then the dimensions of both text and image features are reduced to $d$ (cf. Section 2.2) by using 1$\times$1 convolutions. The decoding module consists of several upsampling layers that transform the learned representations into 64$\times$64 images. For the discriminator, we first apply convolutional layers to encode the images into feature representations. We then concatenate the image representation with text embeddings, then apply two convolutional layers to produce final probabilities. Note that we use concatenation to conditioning for limiting the discriminator capability to prevent the mode collapse effect.

To train the generator and discriminator, we adopt the Adam optimizer with momentum of 0.5. The learning rate is 0.0002. We set batch size to 64 for all three experiments and number of iterative epochs to 600 for birds and flowers synthesis, 200 for fashion synthesis. The parameters of VGG part were fixed during training the generator. The training takes about 1 day to converge on a single Tesla P100 GPU.

\subsection{Qualitative comparison}
We compare our proposed model with the baseline \cite{dong2017semantic} (i.e., concatenation) and FiLM on three datasets. The results are  shown in Fig \ref{fig:exp}. Baseline method fails to transform the detail attributes based on the given description because the learned representations are not powerful as it does not contain enough detail information. For example, the generated images by baseline method in editing flowers demonstrate the model has learned the colors of yellow and orange, but it is unaware of the location of these colors.
Meanwhile, when original image has a complex background (e.g. 3th and 4th samples in first row), the model will fall into mode collapse and output the same meaningless image. On the contrary, our method can capture the specific semantic changes in detail, which is attribute to our richer bilinear representations. It correctly disentangles semantically related objects from some messy images and prevent the occurrence of mode collapse. As a consequence, our approach can successfully generate meaningful images subject to the text description.

\subsection{Quantitative comparison}
We choose inception score (IS) for quantitative evaluation. Inception score is a well-known metric for evaluating GANs. IS can be computed by $IS=\exp(\mathbb{E}_{x}D_{KL}(p(y|x)\| p(y)))$, where $x$ denotes one generated sample, and $y$ is the label predicted by the inception model. The better models which generate diverse and meaningful images can get larger inception score. In this experiment, we use the test dataset for evaluation. We first finetune the inception model with test images for classification. Then, for every test class, we randomly choose an image and text description (e.g. if test dataset has 40 classes, 40 images and 40 descriptions are selected). The images are generated by inputting every pair of images and descriptions (e.g. 40 images and 40 descriptions can generate 40$\times$40 edited images).

The results are shown in Table~\ref{tabel:IS}. 
To explore the influence of rank constraint $d$, we set $d=2,64,256$ and get three variants: Bil-R2, Bil-R64 and Bil-R256. The Bil-R256 gets the highest IS in all three tasks. Interestingly, the baseline method has higher IS than FiLM on Oxford flower dataset because flower editing is simple and is not very dependent on the power of learned representation. For more complicated bird and fashion editings, our method gets the highest IS and achieves better performance with the increasing of $d$. Experimental result suggests that the learned bilinear representation is more powerful and do help to generate images with higher quality.

\begin{table}
\begin{center}
\begin{tabular}{cccc}
\hline
Methods  & Caltech bird & Oxford flower & Fashion \\ \hline
Baseline     &   1.92$\pm$0.05  &    5.03$\pm$0.62 &    8.65$\pm$1.33  \\ 
FiLM \cite{gunel2018language}    &   2.59$\pm$0.11  &    4.83$\pm$0.48 &    8.78$\pm$1.43  \\ 
\textbf{Bil-R2} &    2.60$\pm$0.11 &    4.93$\pm$0.39 &    9.30$\pm$1.48 \\ 
\textbf{Bil-R64}    &    2.63$\pm$0.17 & 5.40$\pm$0.62 & 10.94$\pm$2.28 \\ 
\textbf{Bil-R256}  & \textbf{2.76$\pm$0.08} &    \textbf{6.26$\pm$0.44} &   \textbf{11.63$\pm$2.15} \\  \hline
\end{tabular}
\end{center}
\vspace{-0.8em}
\caption{The comparison of IS score of methods}
\vspace{-1em}
\label{tabel:IS}
\end{table}

\vspace{-3mm}
\section{Conclusion}
\label{sec:conclusion}
In this work, we propose a conditional GAN based encoder-decoder architecture to semantically manipulate images by text descriptions. A general condition layer called Bilinear Residual Layer (BRL) is proposed to learn more powerful bilinear representations for LBIE. BRL is also applicable for other common conditional tasks. Our evaluation results on Caltech-200 bird dataset, Oxford-102 flower dataset and Fashion Synthesis dataset achieve plausible effects and outperform the state-of-art methods on LBIE.

\bibliographystyle{IEEEbib}
\bibliography{strings,refs}

\end{document}